\documentclass{article}

\usepackage[preprint]{neurips_2026}

\usepackage[utf8]{inputenc} 
\usepackage[T1]{fontenc}    
\usepackage{hyperref}       
\usepackage{url}            
\usepackage{booktabs}       
\usepackage{amsfonts}       
\usepackage{nicefrac}       
\usepackage{microtype}      
\usepackage{xcolor}         
\usepackage{multirow}
\usepackage{graphicx}
\usepackage{amsmath}
\usepackage{amssymb}
\usepackage{algorithm}
\usepackage{algpseudocode}
\usepackage[table]{xcolor} 
\usepackage{bm} 
\usepackage{placeins}

\definecolor{highlightgray}{gray}{0.92}
\newcommand{\best}[1]{\cellcolor{highlightgray}$\bm{#1}$}

\title{Data-Asymmetric Latent Imagination and Reranking for 3D Robotic Imitation Learning}

\author{%
\begin{tabular}{c}
\begin{tabular}{cccc}
Lianghao Luo & Xizhou Bu & Ruyan Liu & Qingqiu Huang \\
Chufeng Tang & Xiaoshuai Hao & Hongbo Wang & Wei Li
\end{tabular}
\\[0.6em]
\texttt{\{25213070026, xzbu24, 24210860011\}@m.fudan.edu.cn}
\\
\texttt{\{draco.huang, felix.tang\}@morphi.com}
\\
\texttt{haoxiaoshuai714@163.com \quad Wanghongbo@fudan.edu.cn \quad fd\_liwei@fudan.edu.cn}
\end{tabular}
}

\begin{document}

\maketitle

\begin{abstract}
Robotic imitation learning typically assumes access to optimal demonstrations, yet real-world data collection often yields suboptimal, exploratory, or even failed trajectories. Discarding such data wastes valuable information about environment dynamics and failure modes, which can instead be leveraged to improve decision-making. While 3D policies reduce reliance on high-quality demonstrations through strong spatial generalization, they still require large-scale data to achieve high task success. To address this, we propose DALI-R, a Data-Asymmetric Latent Imagination and Reranking framework for 3D robotic imitation learning from mixed-quality trajectories. It learns a Latent World Model over 3D point clouds for imagined rollouts and a Task Completion Scorer that reranks candidate action chunks, improving decision-making without additional high-quality demonstrations. We instantiate DALI-R with both diffusion and efficient flow-matching policies and evaluate it on Adroit and MetaWorld benchmarks. Across the two evaluated 3D base policies, DALI-R achieves an average  $6.8$\% improvement in success rate while incurring less than  $0.7\times$ additional inference overhead.
\end{abstract}

\section{Introduction}

Imitation learning has become a widely used paradigm for learning complex robotic manipulation skills from demonstrations~\citep{argall2009survey,florence2022implicit,chi2023diffusion}. Recent generative policies, including diffusion-based action models, can represent multimodal action distributions and generate temporally coherent action chunks for high-dimensional control~\citep{chi2023diffusion,zhao2023learning}. These advances have enabled robots to acquire contact-rich manipulation behaviors without manually designing dense reward functions. However, their performance still depends heavily on clean and successful expert demonstrations. In practical robotic data collection, high-quality demonstrations are expensive to obtain, while many collected trajectories are suboptimal, exploratory, or failed. Naively training a behavior cloning policy on mixed-quality trajectories can corrupt the expert action prior~\citep{ross2011dagger,brown2019extrapolating}, whereas discarding imperfect data wastes useful information about environment dynamics, failure modes, and task boundaries.

This tension motivates the central question of this work: how can a robot learning system exploit abundant imperfect data without degrading the policy that imitates expert behavior? Our key observation is that different components of a robot learning system need not consume data of the same quality. A generative policy should imitate successful expert behavior and therefore benefits from clean demonstrations. In contrast, predictive and evaluative models can benefit from broader offline data because suboptimal and failed trajectories reveal how the environment evolves under diverse actions and which states are unlikely to lead to task completion. This suggests a data-asymmetric learning strategy: use expert data to train the action prior, and use mixed-quality data to train models for inference-time prediction and action selection.

A natural way to use such predictive and evaluative models is inference-time
imagination and reranking. Given multiple candidate action chunks from a policy,
a world model can imagine their future consequences, and a task-completion
scorer can select the most promising candidate before execution. Existing predictive-reranking and video-model-based policies often reason in image or
video space~\citep{qi2026gpc,kim2026cosmospolicy,zhang2025worldinworld}. While promising for high-level planning, these pipelines can be computationally expensive for high-frequency robot control and may not explicitly preserve the 3D geometric structure needed for precision manipulation. Since manipulation success often depends on object pose, end-effector alignment, and contact geometry, we instead perform prediction and evaluation in a compact 3D latent representation extracted from point-cloud observations and robot states.

We instantiate this idea as DALI-R, a data-asymmetric latent imagination and
reranking framework for 3D robotic imitation learning. The Base 3D Policy is
trained only on successful expert demonstrations, preserving a clean action
prior over successful behaviors. In parallel, a Latent World Model and a Task
Completion Scorer are trained using mixed-quality trajectories, including
successful expert data, imperfect but successful data, and failed data. The
world model predicts action-conditioned latent state transitions, while the
Task Completion Scorer estimates the probability of task completion from latent
states. Imperfect and failed trajectories are therefore not used as direct
imitation targets; instead, they provide supervision for dynamics prediction and
failure-aware scoring.

At deployment time, the frozen Base 3D Policy proposes multiple candidate action
chunks from the current observation. To obtain diverse candidates without
directly injecting unstructured noise into action space, we apply stochastic
point dropout to the input point cloud and query the policy under these
perturbed observations. Each candidate is then imagined by the Latent World
Model in a compact 3D latent space and evaluated by the Task Completion Scorer.
The robot executes the highest-scoring candidate in a receding-horizon manner.
This inference-time imagination and reranking procedure allows the system to
use information from imperfect and failed data while keeping the Base 3D Policy
trained only on successful expert demonstrations.

Efficiency is critical for making this loop practical. Standard diffusion
policies often require many denoising steps, limiting the number of candidates
that can be generated and evaluated within a control cycle. We therefore also
instantiate the Base 3D Policy using optimal-transport flow matching~\citep{lipman2022flow}, which
learns a deterministic transport vector field from a simple noise distribution
to the expert action distribution. In practice, this enables high-throughput
candidate generation with only a small number of network function evaluations,
making it suitable for latent imagination and scorer-based reranking.

We evaluate our framework on challenging manipulation tasks from Adroit~\citep{rajeswaran2018learning} and
MetaWorld~\citep{yu2020metaworld}. In our main multi-seed evaluation, DALI-R improves the average
success rates of both 3D diffusion and 3D flow-matching policies. Additional
diagnostic ablations suggest that stochastic point dropout, sparse binary
task-completion supervision, and mixed-quality dynamics data each contribute to
its effectiveness. We further conduct an offline real-world diagnostic on point-cloud trajectories
released by 3D Diffusion Policy~\citep{ze20243d}, where action conditioning
substantially reduces held-out latent transition error compared with an
action-ablated world model.

Our contributions are summarized as follows:
\begin{itemize}
    \item We propose DALI-R, a data-asymmetric latent imagination and reranking framework for robotic imitation learning, where successful expert data trains a clean Base 3D Policy, while mixed-quality trajectories train a Latent World Model and Task Completion Scorer for inference-time action selection.

    \item We introduce an inference-time pipeline that combines stochastic
    point dropout, chunk-level latent world-model prediction, and
    task-completion scoring to rerank candidate action chunks without additional online interaction.

    \item We instantiate DALI-R with diffusion and optimal-transport flow-matching backbones, evaluate it on Adroit and MetaWorld, and provide an offline real-world diagnostic on released 3D Diffusion Policy trajectories.
\end{itemize}
\section{Related work}

\paragraph{3D imitation learning and generative visuomotor policies.}
Imitation learning is a standard paradigm for acquiring robotic manipulation skills from demonstrations without manually designing dense rewards~\citep{argall2009survey,florence2022implicit}. Early behavior cloning methods often learn deterministic or unimodal action mappings, which can be insufficient for contact-rich manipulation where multiple plausible actions may exist under the same observation. Recent generative visuomotor policies address this issue by modeling multimodal action distributions and generating temporally coherent action chunks~\citep{zhao2023learning,chi2023diffusion}. Building on denoising diffusion models and deterministic diffusion sampling~\citep{ho2020denoising,song2020denoising}, Diffusion Policy~\citep{chi2023diffusion} formulates action prediction as conditional denoising and has shown strong performance in manipulation tasks. Follow-up works incorporate 3D or multi-view observations to better preserve object geometry, end-effector pose, and contact structure~\citep{shridhar2023perceiver,gervet2023act3d,goyal2023rvt,ze20243d}. Recent flow-based and consistency-based generative models further improve inference efficiency for 3D manipulation~\citep{lipman2022flow,song2023consistency}; for example, FlowPolicy applies consistency flow matching to point-cloud-conditioned policy generation and reduces inference latency while maintaining competitive success rates~\citep{zhang2025flowpolicy}. These methods provide expressive and efficient action priors for high-dimensional control, but they are typically trained by cloning successful demonstrations. In contrast, DALI-R is complementary to the choice of action prior: it keeps the 3D generative policy trained only on expert demonstrations, and uses imperfect trajectories to train separate predictive and evaluative models for inference-time action selection.

\paragraph{Learning from imperfect demonstrations and mixed-quality data.}
Robotic data collection often produces trajectories of varying quality, including suboptimal, exploratory, or failed executions. A large body of work studies how to learn from such imperfect data through offline reinforcement learning~\citep{kumar2020conservative,kostrikov2021offline}, inverse reinforcement learning, or reward
learning from suboptimal demonstrations~\citep{ziebart2008maximum,brown2019extrapolating}. These methods aim to extract useful supervision from non-expert data, but directly incorporating imperfect trajectories into policy learning can be risky when the action prior is expected to imitate high-quality behavior. Behavior cloning on mixed-quality data may pull the learned policy toward unsuccessful or inefficient actions, while purely filtering demonstrations may discard informative failure cases that reveal task boundaries and transition dynamics. DALI-R follows a different design principle: data quality should be matched to model role. Expert demonstrations train the action prior, whereas mixed-quality trajectories train the Latent World Model and Task Completion Scorer. Thus, imperfect data contributes to prediction and evaluation rather than direct action supervision.

\paragraph{Inference-time planning, reranking, and world models.}
Another line of work improves decision making at inference time by generating multiple candidate actions or trajectories and selecting among them using a learned model, reward function, value function, or simulator. Classical model-predictive control and trajectory optimization methods, such as random shooting, CEM~\citep{deboer2005tutorial}, and MPPI~\citep{williams2017information}, perform online search in action space. Earlier visual foresight methods plan through learned image-space prediction~\citep{finn2017deep}.
Recent learning-based approaches combine generative policies with value-guided sampling, diffusion-based planning, or learned world models to evaluate candidate futures before execution~\citep{janner2022planning,hansen2022tdmpc,hafner2023dreamerv3}. Closest to our work is Generative Predictive Control (GPC), which enhances a frozen diffusion behavior-cloning policy at inference time by using an action-conditioned predictive world model to rank or refine action proposals~\citep{qi2026gpc}. Recent video-model-based policies also integrate action generation, future-state prediction, and value estimation~\citep{du2023video,kim2026cosmospolicy}; for example, Cosmos Policy adapts a pretrained video model to generate action chunks, future states, and values for best-of-$N$ planning. Broader closed-loop benchmarks such as World-in-World further highlight that world models should be evaluated by embodied task success rather than visual fidelity alone~\citep{zhang2025worldinworld}. DALI-R differs from these image- or video-space approaches by focusing on 3D robotic imitation from mixed-quality demonstrations: it keeps the base 3D policy frozen and expert-trained, uses suboptimal and failed trajectories only for latent prediction and success evaluation, and performs lightweight reranking in a compact 3D latent space.
\begin{figure}[t]
    \centering
    \includegraphics[width=\textwidth]{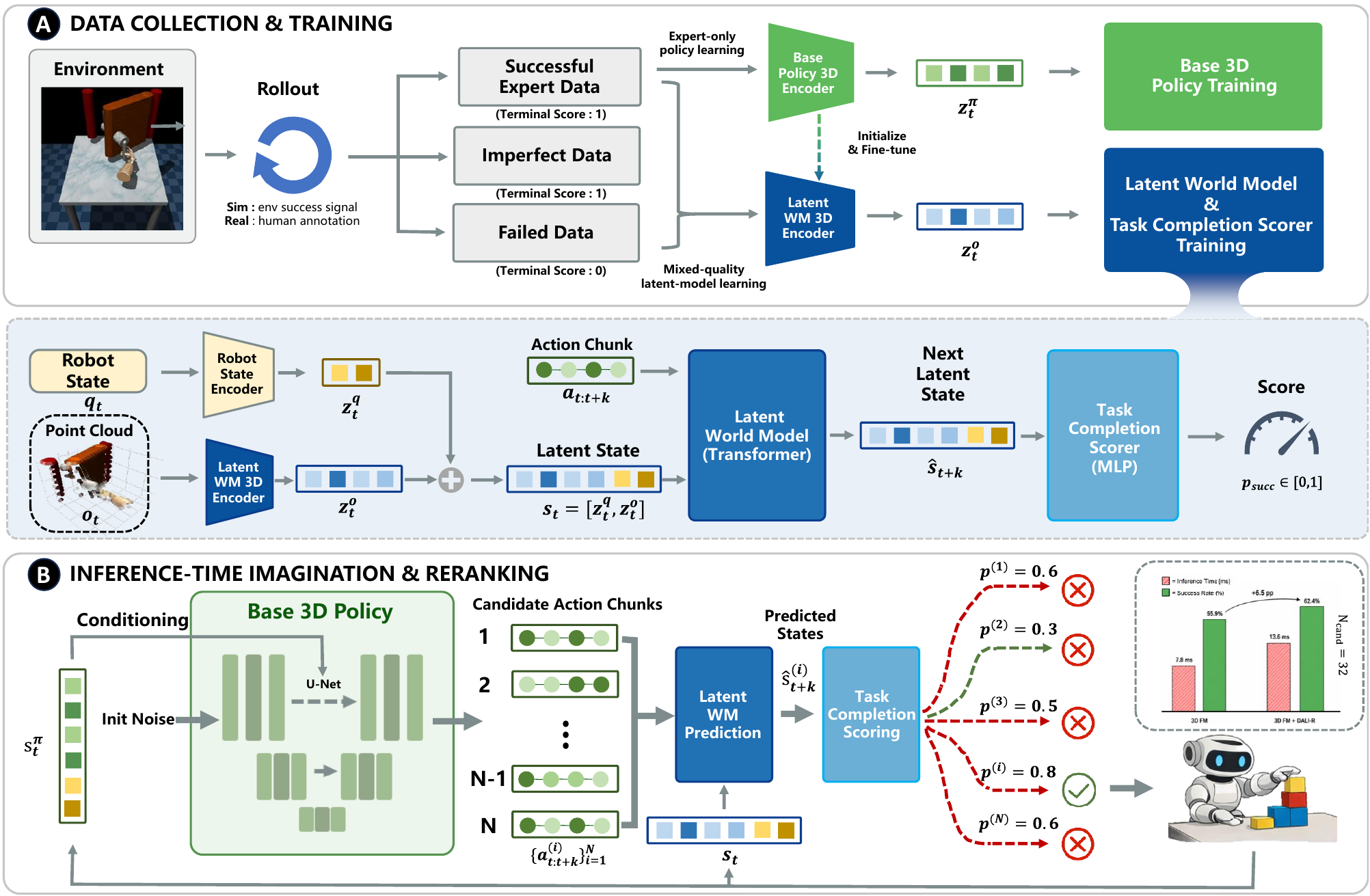}
    \vspace{-0.3cm}
    \caption{
    \textbf{Overview of DALI-R, our data-asymmetric latent imagination and reranking framework.}
    \textbf{(A)} During training, the Base 3D Policy is trained only on successful expert data, while mixed-quality trajectories, including imperfect-success and failed data, are used to train the Latent World Model and Task Completion Scorer.
    \textbf{(B)} At inference time, stochastic point dropout produces perturbed point-cloud observations for candidate generation. Candidate action chunks from the frozen Base 3D Policy are imagined in latent space and scored by the Task Completion Scorer, and the highest-scoring candidate is executed in a receding-horizon loop.
    }
    \label{fig:system_overview}
\end{figure}

\section{Methodology}
\label{sec:method}

\subsection{Problem setup and data-asymmetric design}

We study visuomotor robotic manipulation under partial observability. At each time step $t$, the robot observes a point cloud $o_t$ and a robot state $q_t$, and predicts a continuous action chunk $\mathbf{a}_{t:t+k}$ for closed-loop control, where $k$ is the action horizon. The policy is learned from offline demonstrations without dense task rewards during imitation learning.

DALI-R targets the common setting where collected trajectories have uneven
quality. We use $\mathcal{D}_{\mathrm{exp}}$ to denote successful expert
demonstrations, $\mathcal{D}_{\mathrm{imp}}^{+}$ to denote imperfect but
successful trajectories, and $\mathcal{D}_{\mathrm{fail}}^{0}$ to denote failed trajectories. The superscripts indicate terminal task-completion labels, where successful trajectories have terminal score $1$ and failed trajectories have terminal score $0$. The mixed-quality dataset is
$\mathcal{D}_{\mathrm{mix}}=\mathcal{D}_{\mathrm{exp}}\cup
\mathcal{D}_{\mathrm{imp}}^{+}\cup\mathcal{D}_{\mathrm{fail}}^{0}$. The central design principle is \emph{data-asymmetric learning}: different model components consume data of different quality according to their roles. The Base 3D Policy is trained only on successful expert demonstrations, while mixed-quality trajectories are used to train latent prediction and task-completion scoring models.

We write the Base 3D Policy as $\pi_\theta=(E^\pi_\theta,G_\theta)$, where
$E^\pi_\theta$ denotes the policy-side observation encoder and $G_\theta$ is the generative action head. In implementation, $E^\pi_\theta$ encodes both the point cloud $o_t$ and the robot state $q_t$; Figure~\ref{fig:system_overview} visualizes its 3D point-cloud branch for compactness. Let $z_t^\pi$ denote the policy-side latent used internally by $\pi_\theta$, distinct from the latent state used by the Latent World Model. After expert-only policy training, we initialize a separate Latent WM 3D Encoder $E^o_\phi$ from the trained policy point-cloud encoder. The copied encoder $E^o_\phi$ is then fine-tuned jointly with the Robot State Encoder $E^q_\eta$, the Latent World Model $W_\psi$, and the Task Completion Scorer $C_\omega$ on $\mathcal{D}_{\mathrm{mix}}$, while the Base 3D Policy $\pi_\theta$ remains frozen.

\begin{figure}[t]
    \centering
    \includegraphics[width=\linewidth]{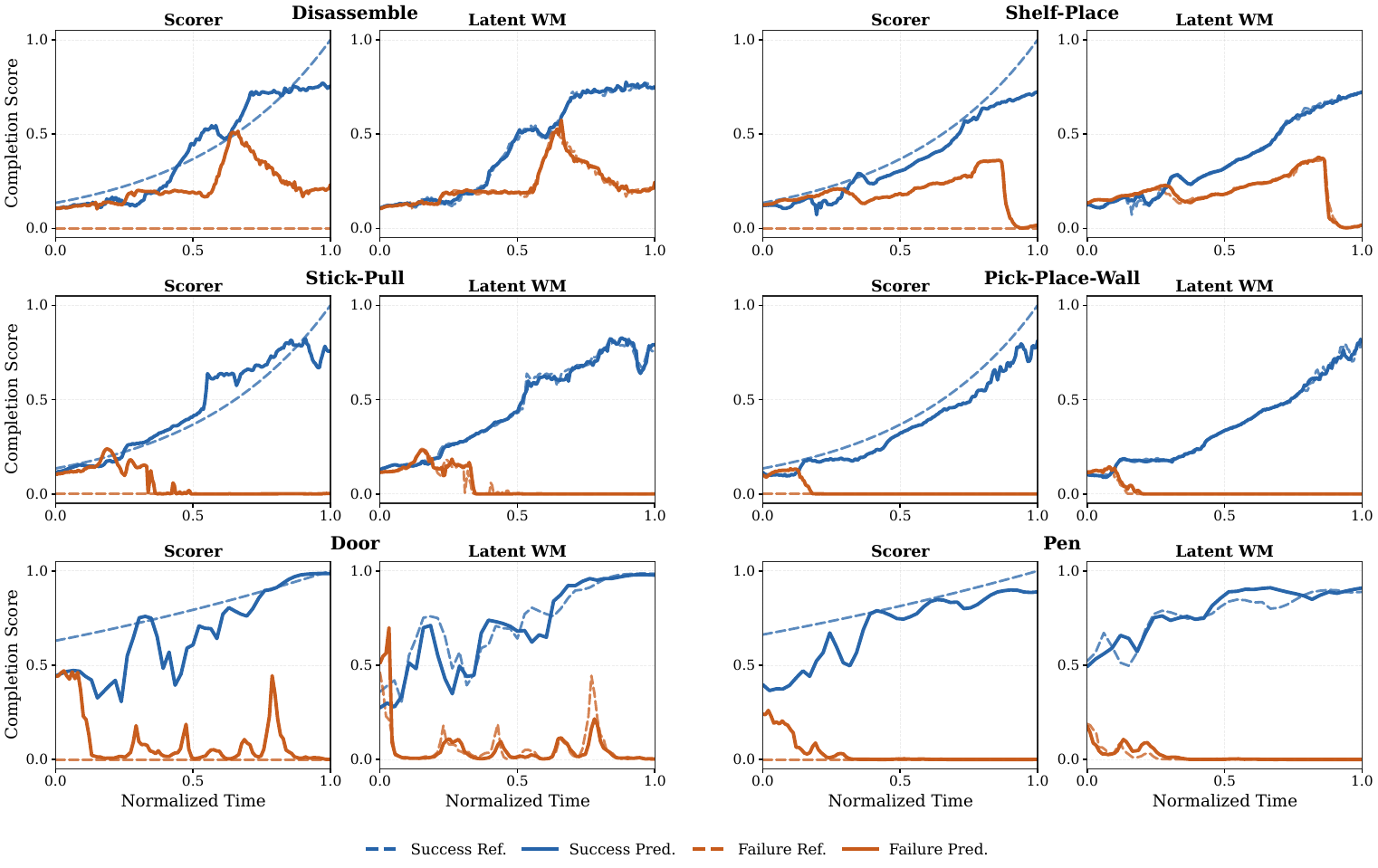}
    \vspace{-0.5cm}
    \caption{
    \textbf{Diagnostic visualization of the learned Task Completion Scorer and Latent World Model.} Scorer plots compare predicted completion scores with Monte-Carlo targets, while WM plots compare scores from predicted and ground-truth future latents. Dashed/solid curves denote references/predictions, and blue/orange curves denote successful/failed trajectories.
    }
    \label{fig:vf_wm_diagnostics}
\end{figure}

\subsection{Expert-trained base 3D policy}

The Base 3D Policy maps the current point cloud $o_t$ and robot state $q_t$ to
a temporally coherent action chunk $\mathbf{a}_{t:t+k}$. Since the policy is trained exclusively on $\mathcal{D}_{\mathrm{exp}}$, it learns an action prior concentrated around successful expert behavior rather than directly cloning suboptimal or failed actions.

DALI-R is agnostic to the specific generative action backbone.
In our experiments, we instantiate the Base 3D Policy with both
diffusion-based 3D action generation~\citep{chi2023diffusion,ze20243d}
and an optimal-transport flow-matching variant~\citep{lipman2022flow,zhang2025flowpolicy}. For the flow-matching variant, the model learns a conditional vector field that transports Gaussian noise to expert action chunks under the current point-cloud observation and robot state. At inference, an action chunk is generated by integrating this vector field with a small number of network function evaluations. This efficient generation process is important because DALI-R samples multiple candidates at each control step.

\subsection{Latent imagination from mixed-quality data}

DALI-R evaluates candidate actions in a compact 3D latent space rather than in image or video space. Given the current point cloud and robot state, the Latent WM 3D Encoder and Robot State Encoder produce $z_t^o=E^o_\phi(o_t)$ and $z_t^q=E^q_\eta(q_t)$, respectively. We concatenate them into the latent state $s_t=[z_t^q,z_t^o]$. For each training segment, the target future latent state $s_{t+k}$ is constructed analogously from the future point cloud and robot state. The Latent World Model $W_\psi$ predicts a chunk-level residual transition from the current latent state and an action chunk, written as $\hat{s}_{t+k}=s_t+W_\psi(s_t,\mathbf{a}_{t:t+k})$, and is trained with a supervised latent prediction loss on $\mathcal{D}_{\mathrm{mix}}$.

The Task Completion Scorer $C_\omega$ estimates the probability that a latent state will eventually lead to task completion. We assume only sparse terminal success labels, with successful and failed trajectories assigned terminal labels $1$ and $0$, respectively. For a trajectory with terminal label $y_T\in\{0,1\}$, we define discounted Monte-Carlo targets
$y_t=\gamma^{T-t}y_T$ and train $C_\omega$ with binary cross entropy on
$p_{\mathrm{succ}}(s_t)=\sigma(C_\omega(s_t))$, where $T$ is the terminal step, $\gamma\in(0,1]$ is the discount factor, and $\sigma$ is the sigmoid function. Thus, mixed-quality data provides supervision about both transition structure and failure modes, while never serving as direct imitation targets for the frozen policy.

Together, $W_\psi$ and $C_\omega$ form a latent imagination module. The world model predicts the latent consequence of a candidate action chunk, and the Task Completion Scorer evaluates whether the imagined outcome is likely to lead to success. Figure~\ref{fig:vf_wm_diagnostics} provides trajectory-level diagnostics showing that the scorer tracks Monte-Carlo task-completion targets and that world-model predictions preserve the success/failure score trends in latent space.

\subsection{Inference-time latent reranking}

At deployment, DALI-R improves the frozen expert-trained Base 3D Policy through best-of-$N_{\mathrm{cand}}$ latent reranking, where $N_{\mathrm{cand}}$ denotes the number of candidate action chunks. To generate diverse candidates without injecting unstructured noise into action space, we apply stochastic point dropout to the input point cloud. For each candidate $i$, point dropout produces a perturbed point cloud $\tilde{o}^{(i)}_t$, and the frozen policy samples an action chunk
$\mathbf{a}^{(i)}_{t:t+k}\sim\pi_\theta(\cdot\mid\tilde{o}^{(i)}_t,q_t)$. Because diversity is induced in observation space, the resulting candidates remain constrained by the expert-trained generative prior rather than drifting into arbitrary action-space perturbations.

The unperturbed point cloud and robot state are encoded by the fine-tuned Latent WM 3D Encoder and Robot State Encoder to form $s_t=[z_t^q,z_t^o]$. Each candidate is then imagined once in latent space and scored by the Task
Completion Scorer. DALI-R selects

\begin{equation}
    i^\star
    =
    \arg\max_{i\in\{1,\ldots,N_{\mathrm{cand}}\}}
    \sigma
    \left(
        C_\omega
        \left(
            s_t + W_\psi(s_t,\mathbf{a}^{(i)}_{t:t+k})
        \right)
    \right),
    \qquad
    \mathbf{a}^{\star}_{t:t+k}=\mathbf{a}^{(i^\star)}_{t:t+k}.
    \label{eq:latent_reranking}
\end{equation}

The robot executes $\mathbf{a}^{\star}_{t:t+k}$ according to the
receding-horizon control schedule and repeats the procedure at the next
observation. Since both prediction and scoring are performed in latent space, the reranking loop remains lightweight and avoids expensive image- or video-space rollout, making it suitable for high-throughput action proposal backbones such as flow matching.
\section{Experiments}
\label{sec:experiments}

We evaluate DALI-R on Adroit~\citep{rajeswaran2018learning} and
MetaWorld~\citep{yu2020metaworld}, two MuJoCo-based manipulation
benchmark suites~\citep{todorov2012mujoco}, covering main success rates,
inference-time action improvement, efficiency, scorer supervision,
mixed-quality latent dynamics, and an offline real-world diagnostic. We
abbreviate 3D Diffusion Policy and 3D Flow Matching Policy as 3D DP and 3D FM, respectively. Main results use three seeds with 100 episodes per task; unless otherwise noted, ablations use one seed with 100 episodes and are intended as diagnostic trends. The real-world diagnostic uses released 3D Diffusion Policy point-cloud trajectories and evaluates offline latent transition prediction rather than closed-loop robot performance.

\begin{table}[t]
\centering
\caption{
Success rates across Adroit and MetaWorld benchmarks.
Results are reported as mean $\pm$ standard deviation over three random seeds, with 100 evaluation episodes per task.
Bold entries with gray backgrounds indicate the best performance in each task.
}
\label{tab:main_results}
\resizebox{\textwidth}{!}{%
\begin{tabular}{l cc cccc c}
\toprule
\multirow{2}{*}{\textbf{Method}} & \multicolumn{2}{c}{\textbf{Adroit (Dexterous Hand)}} & \multicolumn{4}{c}{\textbf{MetaWorld (General Gripper)}} & \multirow{2}{*}{\textbf{Total Avg.}} \\
\cmidrule(lr){2-3} \cmidrule(lr){4-7} 
& Door & Pen & Disassemble & Shelf Place &  Stick Pull & Pick Place Wall & \\
\midrule
\textit{2D Baselines} \\
Diffusion Policy~\citep{chi2023diffusion}     & $42.7 \pm 2.4$ & $31.7 \pm 4.7$ & $56.3 \pm 4.2$ & $32.7 \pm 5.0$ & $48.0 \pm 2.1$ & $54.3 \pm 2.6$ & $44.3$ \\
\midrule
\textit{3D Base Policies} \\
3D Diffusion Policy~\citep{ze20243d}       & $50.7 \pm 2.7$ & $46.2 \pm 4.9$ & $67.7 \pm 2.5$ & $46.7 \pm 3.9$ & $49.3 \pm 1.7$ & $61.7 \pm 1.2$ & 53.7 \\
3D Flow Matching Policy~\citep{zhang2025flowpolicy}   & $47.2 \pm 1.4$ & $50.8 \pm 2.2$ & $76.3 \pm 2.9$ & $49.3 \pm 2.1$ & $48.7 \pm 1.2$ & $63.3 \pm 2.5$ & 55.9 \\
\midrule
\textit{Ours (3D Reranking)} \\
\textbf{3D DP + Ours} & \best{57.8 \pm 2.1} & $52.0 \pm 3.5$ & $79.0 \pm 3.7$ & $53.3 \pm 2.9$ & \best{53.0 \pm 2.8} & $69.3 \pm 0.9$ & 60.7 \\
\textbf{3D FM + Ours} & $54.8 \pm 2.7$ & \best{56.3 \pm 0.8} & \best{84.7 \pm 1.2} & \best{56.3 \pm 1.7} & $50.7 \pm 1.7$ & \best{71.3 \pm 2.1} & \best{62.4} \\
\bottomrule
\end{tabular}%
}
\end{table}

\subsection{Main results: latent reranking improves 3D generative policies}
\label{sec:main_results}

Table~\ref{tab:main_results} reports the main multi-seed success rates across
Adroit and MetaWorld tasks.
Compared with the 2D diffusion baseline, both 3D base policies achieve higher
average success rates, highlighting the importance of preserving 3D geometric
structure for manipulation.
The 3D Diffusion Policy obtains an average success rate of $53.7\%$, while the
3D Flow Matching Policy reaches $55.9\%$.

Adding DALI-R consistently improves both 3D policy backbones. For 3D Diffusion Policy, our method increases the average success rate from $53.7\%$ to $60.7\%$, corresponding to a $7.0\%$ absolute improvement. For 3D Flow Matching, our method improves the average success rate from $55.9\%$ to $62.4\%$, a $6.5\%$ absolute gain. These improvements are observed across all evaluated tasks, suggesting that the learned Latent World Model and Task Completion Scorer provide useful inference-time action selection beyond the base generative prior.

The gains are especially visible on tasks that require precise pose control,
contact reasoning, and spatial alignment. For example, with DALI-R, 3D Flow
Matching improves from $76.3\%$ to $84.7\%$ on MetaWorld Disassemble and from
$50.8\%$ to $56.3\%$ on Adroit Pen. These results support the central design of
DALI-R: imperfect trajectories are not used as imitation targets, but instead
provide transition and failure-mode supervision for inference-time latent
reranking.

\subsection{Using latent imagination for inference-time action improvement}
\label{sec:obs_candidate}

A key question is how to use the learned Latent World Model and Task Completion Scorer to improve actions at inference time. A naive strategy is to repeatedly query the frozen imitation policy and rerank the resulting action chunks. However, as visualized in Figure~\ref{fig:efficiency_scaling}(c), clean policy samples tend to be low-diversity and concentrated around similar actions, providing limited candidate variation for reranking. We therefore compare two broader ways of introducing candidate diversity: varying the action chunk itself through scorer-guided search or optimization, and varying the policy input through observation perturbations while keeping the expert-trained policy as the action proposal mechanism. Table~\ref{tab:action_improvement} compares these alternatives, including random shooting, CEM~\citep{deboer2005tutorial}, MPPI~\citep{williams2017information}, and score-gradient ascent, using the same 3D FM base policy and Task Completion Scorer.

Action-chunk variation provides limited and inconsistent improvements over the base policy. In our experiments, methods that vary the action chunk directly often reduce the average success rate relative to the unrefined base policy. This suggests that scorer-guided search or optimization in action space can move candidates away from the expert action manifold, where the learned scorer may be less reliable.

Varying the policy input is more effective. Standard observation noise improves the average success rate from $56.6\%$ to $60.9\%$, while stochastic point dropout further improves it to $63.8\%$. Unlike direct action-chunk variation, point dropout perturbs the policy input while leaving the expert-trained generative prior responsible for producing temporally coherent action chunks. These results suggest that latent imagination is most effective when used to evaluate and rerank policy-generated candidates, rather than to directly optimize unconstrained action proposals.

\begin{table}[t]
\centering
\caption{
Single-seed diagnostic of inference-time action improvement using the same 3D FM policy and Task Completion Scorer.
All methods select candidate action chunks with the same scorer, but introduce variation either in action space or policy input.
}
\label{tab:action_improvement}
\resizebox{\textwidth}{!}{%
\begin{tabular}{l l cc cccc c}
\toprule
\multirow{2}{*}{\textbf{Method}} & \multirow{2}{*}{\textbf{Variation Target}} & \multicolumn{2}{c}{\textbf{Adroit}} & \multicolumn{4}{c}{\textbf{MetaWorld}} & \multirow{2}{*}{\textbf{Avg.}} \\
\cmidrule(lr){3-4} \cmidrule(lr){5-8}
& & Door & Pen & Disassemble & Shelf Place & Stick Pull & Pick Place Wall & \\
\midrule
Base Policy & None & 47.0 & 49.5 & 76.0 & 52.0 & 49.0 & 66.0 & 56.6 \\
Random Shooting & Action Chunk & 51.0 & \best{61.0} & 73.0 & 30.0 & 49.0 & 56.0 & 53.3 \\
CEM & Action Chunk & 48.0 & 57.0 & 74.0 & 33.0 & \best{52.0} & 59.0 & 53.8 \\
MPPI & Action Chunk & 54.0 & 55.0 & 80.0 & 30.0 & 47.0 & 56.0 & 53.7 \\
Score Grad. Ascent & Action Chunk & 31.0 & 57.0 & 80.0 & 31.0 & 46.0 & 52.0 & 49.5 \\
Std. Obs. Noise & Policy Input & 49.0 & 57.5 & 82.0 & 55.0 & 48.0 & 70.0 & 60.9 \\
Point Dropout & Policy Input & \best{58.0} & 55.5 & \best{85.0} & \best{58.0} & 50.0 & \best{74.0} & \best{63.8} \\
\bottomrule
\end{tabular}%
}
\end{table}

\subsection{Inference efficiency and candidate scaling}
\label{sec:efficiency_scaling}

\begin{figure}[t]
    \centering
    \includegraphics[width=\linewidth]{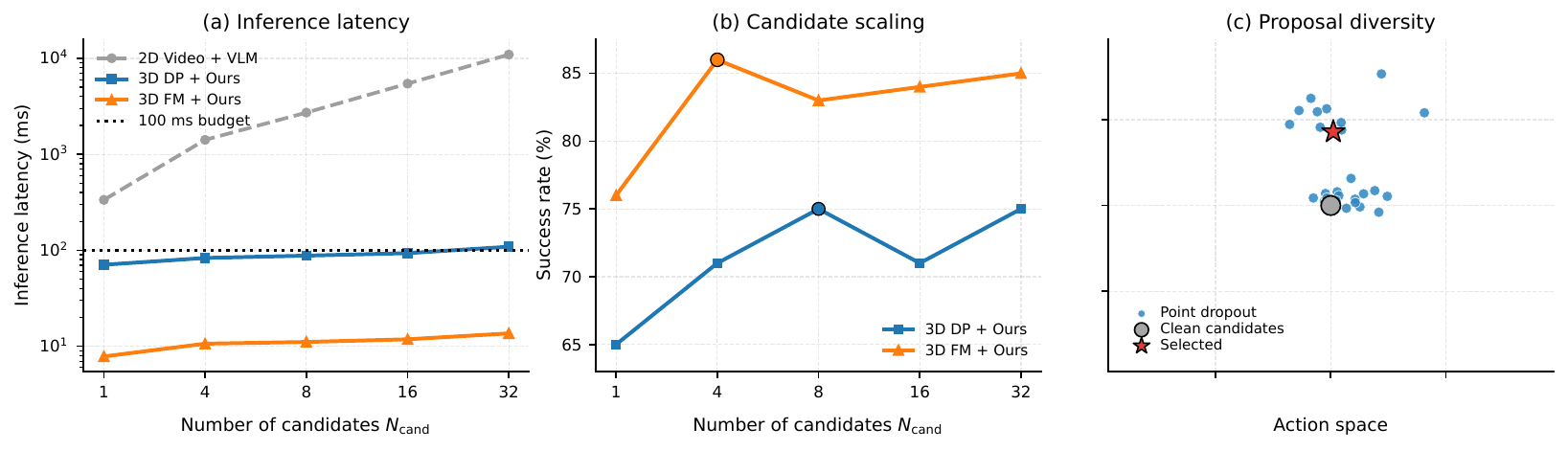}
    \vspace{-0.7cm}
    \caption{
    \textbf{Inference-time efficiency and candidate generation analysis.}
    \textbf{(a)} Latency under different $N_{\mathrm{cand}}$, with 2D Video + VLM shown only as a simulated latency reference and the dotted line denoting a 100 ms budget.
    \textbf{(b)} Single-seed candidate-scaling diagnostic on Disassemble for 3D DP + Ours and 3D FM + Ours.
    \textbf{(c)} Proposal-diversity visualization: clean candidates collapse near one action, while point dropout produces diverse policy-generated candidates for reranking.
    }
    \label{fig:efficiency_scaling}
\end{figure}

We next analyze the latency and candidate-scaling behavior of DALI-R, since reranking must run inside a closed-loop control cycle. Figure~\ref{fig:efficiency_scaling}(a) reports inference latency under different numbers of candidate action chunks $N_{\mathrm{cand}}$.
The 2D Video + Vision-Language Model (VLM) pipeline is included only as a simulated latency reference rather than a trained task baseline, approximating a GPC-style predictive-reranking pipeline~\citep{qi2026gpc} with a lightweight visual-generation proxy and Qwen2-VL-2B scoring~\citep{wang2024qwen2vl}. Its latency reaches $1414.91$ ms at $N_{\mathrm{cand}}=4$ and $10948.59$ ms at $N_{\mathrm{cand}}=32$, making it unsuitable for high-frequency control.

In contrast, 3D latent reranking remains efficient.
At $N_{\mathrm{cand}}=4$, 3D DP + Ours requires $83.17$ ms, while 3D FM + Ours requires only $10.64$ ms. Even at $N_{\mathrm{cand}}=32$, 3D FM + Ours requires only 13.62 ms per decision, corresponding to $73.44$ FPS, showing that flow matching provides a high-throughput action proposal backbone for latent reranking.

Figure~\ref{fig:efficiency_scaling}(b) shows a single-seed candidate-scaling diagnostic on Disassemble for 3D DP + Ours and 3D FM + Ours. Increasing $N_{\mathrm{cand}}$ initially improves success, but the trend is not strictly monotonic: 3D FM + Ours improves from $76.0\%$ at $N_{\mathrm{cand}}=1$ to $86.0\%$ at $N_{\mathrm{cand}}=4$, then fluctuates for larger candidate sets. This is expected because the learned Task Completion Scorer is not an oracle; larger candidate pools may include better actions, but can also introduce overestimated out-of-distribution candidates. Main results use $N_{\mathrm{cand}}=32$, while this diagnostic shows that smaller candidate sets already yield useful gains.

\subsection{Sparse supervision for task-completion scoring}
\label{sec:sparse_supervision}

We next examine the supervision used to train the Task Completion Scorer. In realistic robotic data collection, dense task rewards or privileged simulator states are often unavailable, whereas terminal task-completion labels can be obtained from environment success signals in simulation or human annotation in real settings. Therefore, our default scorer is trained only from sparse
terminal $0/1$ success labels. To contextualize this design choice, Table~\ref{tab:diagnostic_ablations}(a) compares three settings on Pen and Disassemble: the base policy without scorer-based reranking, our scorer trained from sparse terminal success labels, and a privileged oracle scorer trained with
dense simulator rewards.

The sparse scorer improves the average success rate from $62.75\%$ to $70.25\%$,
substantially outperforming the base policy. The dense-reward oracle reaches $73.5\%$, leaving only a $3.25\%$ absolute gap relative to the sparse model. This comparison suggests that terminal task-completion labels provide a practical and effective training signal for scorer-based reranking, while avoiding privileged reward supervision that would not generally be available in real-world data. Thus, DALI-R uses dense rewards only as a diagnostic oracle in this ablation, not
as part of the default training pipeline.

\subsection{Effect of mixed-quality data on latent dynamics}
\label{sec:mixed_data}

We further analyze how the amount of mixed-quality data affects the Latent World Model.
Table~\ref{tab:diagnostic_ablations}(b) reports latent prediction MSE and downstream success rate on the Disassemble diagnostic setting as the number of mixed trajectories increases while the expert data remains fixed.
Increasing the amount of mixed data from $100$ to $1000$ trajectories reduces the latent MSE from $0.539$ to $0.381$ and improves downstream success from $82.0\%$ to $85.0\%$.

This trend supports the data-asymmetric design of DALI-R.
Although imperfect trajectories may be harmful as direct imitation targets, they provide valuable coverage of state transitions and failure modes for learning latent dynamics.
Better latent prediction, in turn, improves the reliability of imagined rollouts used during inference-time reranking.

\begin{table}[t]
\centering
\caption{
Diagnostic ablations for scorer supervision and mixed-quality latent dynamics.
(a) compares sparse terminal labels with dense reward supervision, and (b) evaluates the effect of increasing mixed-quality data for Latent World Model training.
}
\label{tab:diagnostic_ablations}
\small
\setlength{\tabcolsep}{4pt}
\renewcommand{\arraystretch}{1.05}

\begin{minipage}[t]{0.54\linewidth}
\centering
\textbf{(a) Scorer supervision}
\vspace{0.25em}

\begin{tabular*}{\linewidth}{@{\extracolsep{\fill}}lccc@{}}
\toprule
Supervision & Pen & Disassemble & Avg. \\
\midrule
No scorer & 49.5 & 76.0 & 62.75 \\
Sparse terminal labels & 55.5 & 85.0 & 70.25 \\
Dense simulator reward & 61.0 & 86.0 & 73.50 \\
\bottomrule
\end{tabular*}
\end{minipage}
\hfill 
\begin{minipage}[t]{0.42\linewidth}
\centering
\textbf{(b) Mixed-quality data}
\vspace{0.25em}

\begin{tabular*}{\linewidth}{@{\extracolsep{\fill}}lcc@{}}
\toprule
WM Data & MSE $\downarrow$ & Success $\uparrow$ \\
\midrule
100 mixed & 0.539 & 82.0\% \\
500 mixed & 0.417 & 84.0\% \\
1000 mixed & 0.381 & 85.0\% \\
\bottomrule
\end{tabular*}
\end{minipage}

\vspace{-0.5em}
\end{table}

\subsection{Offline validation on real-world point clouds}
\label{sec:real_world}

We additionally evaluate the Latent World Model on real-world point-cloud
trajectories released by 3D Diffusion Policy~\cite{ze20243d}.
We split trajectories at the trajectory level to avoid frame-level leakage and compare the full action-conditioned world model with an action-ablated baseline. Since real-world point clouds often contain large static background regions, we focus on whether action conditioning improves held-out latent transition prediction rather than interpreting raw cross-domain MSE in isolation. This diagnostic tests whether the learned latent dynamics capture action-dependent transitions under real sensor observations, where background clutter, calibration noise, and partial visibility may differ from simulation.

As shown in Table~\ref{tab:real_world_ablation}, action conditioning consistently reduces latent transition error, decreasing the average MSE from $0.086$ to $0.010$ and reducing the \textit{Drill} error from $0.047$ to $0.017$. The improvement is also observed on the other two real-world tasks, suggesting that the benefit is not limited to a single trajectory distribution. These results indicate that action chunks contain transition-relevant information under real sensor observations and that the latent imagination module can model action-dependent transition structure beyond simulation. While this does not replace closed-loop real-robot evaluation, it provides evidence that the proposed latent prediction objective remains meaningful on released real-world point-cloud trajectories.

\begin{table}[t]
\centering
\caption{
Offline latent transition MSE on released 3D Diffusion Policy trajectories; lower is better.
We compare the full action-conditioned Latent World Model with an action-ablated variant.
}
\label{tab:real_world_ablation}
\begin{tabular*}{\linewidth}{@{\extracolsep{\fill}}lcccc@{}}
\toprule
\textbf{Method} & \textbf{Drill (Real World)} & \textbf{Pour (Real World)} & \textbf{Dumpling (Real World)} & \textbf{Avg.} \\
\midrule
WM w/o Action & 0.047 & 0.195 & 0.016 & 0.086 \\
WM w/ Action  & \textbf{0.017} & \textbf{0.010} & \textbf{0.004} & \textbf{0.010} \\
\bottomrule
\end{tabular*}
\end{table}
\section{Conclusion and limitations}

\textbf{Conclusion.}
We presented DALI-R, a data-asymmetric latent imagination and reranking framework for 3D robotic imitation learning from mixed-quality trajectories.
DALI-R separates imitation, prediction, and evaluation: a clean 3D generative policy is trained only from expert demonstrations, while mixed-quality trajectories are used to train a Latent World Model and Task Completion Scorer for inference-time action selection.
This design allows imperfect and failed trajectories to provide transition and failure-mode supervision without corrupting the expert action prior.
Across Adroit and MetaWorld tasks, DALI-R improves both 3D diffusion and 3D flow-matching policies while remaining efficient enough for closed-loop reranking. Our diagnostic analyses further support the roles of observation-space candidate generation, sparse task-completion supervision, and mixed-quality latent dynamics. The offline real-world diagnostic suggests that action-conditioned latent prediction remains useful under real-world point-cloud observations. Overall, these results indicate that imperfect trajectories can be valuable for prediction and evaluation even when they are unsuitable as direct imitation targets.

\textbf{Limitations.}
DALI-R has several methodological limitations. First, it relies on accurate and well-calibrated latent prediction and task-completion scoring; overconfident scores may lead to suboptimal reranking. Second, the current chunk-level reranking procedure provides limited explicit reasoning over long-horizon dependencies and delayed outcomes. Third, closed-loop deployment on physical robots remains necessary to assess robustness under sensor noise, dynamics mismatch, and execution uncertainty. Future work may incorporate uncertainty-aware scoring, multi-step latent rollout, and real-robot validation.

{
\clearpage
\small
\bibliographystyle{unsrt}
\bibliography{references}

@inproceedings{chi2023diffusion,
  author       = {Cheng Chi and
                  Siyuan Feng and
                  Yilun Du and
                  Zhenjia Xu and
                  Eric Cousineau and
                  Benjamin Burchfiel and
                  Shuran Song},
  title        = {Diffusion Policy: Visuomotor Policy Learning via Action Diffusion},
  booktitle    = {Robotics: Science and Systems XIX, Daegu, Republic of Korea, July
                  10-14, 2023},
  year         = {2023},
  url          = {https://doi.org/10.15607/RSS.2023.XIX.026},
  doi          = {10.15607/RSS.2023.XIX.026}
}

@inproceedings{ze20243d,
  author       = {Yanjie Ze and
                  Gu Zhang and
                  Kangning Zhang and
                  Chenyuan Hu and
                  Muhan Wang and
                  Huazhe Xu},
  title        = {3D Diffusion Policy: Generalizable Visuomotor Policy Learning via
                  Simple 3D Representations},
  booktitle    = {Robotics: Science and Systems XX, Delft, The Netherlands, July 15-19,
                  2024},
  year         = {2024},
  url          = {https://doi.org/10.15607/RSS.2024.XX.067},
  doi          = {10.15607/RSS.2024.XX.067}
}

@inproceedings{kumar2020conservative,
  author       = {Aviral Kumar and
                  Aurick Zhou and
                  George Tucker and
                  Sergey Levine},
  title        = {Conservative Q-Learning for Offline Reinforcement Learning},
  booktitle    = {Advances in Neural Information Processing Systems 33: Annual Conference
                  on Neural Information Processing Systems 2020, NeurIPS 2020, December
                  6-12, 2020, virtual},
  year         = {2020},
  url          = {https://proceedings.neurips.cc/paper/2020/hash/0d2b2061826a5df3221116a5085a6052-Abstract.html}
}

@inproceedings{brown2019extrapolating,
  author       = {Daniel S. Brown and
                  Wonjoon Goo and
                  Prabhat Nagarajan and
                  Scott Niekum},
  title        = {Extrapolating Beyond Suboptimal Demonstrations via Inverse Reinforcement
                  Learning from Observations},
  booktitle    = {Proceedings of the 36th International Conference on Machine Learning,
                  {ICML} 2019, 9-15 June 2019, Long Beach, California, {USA}},
  series       = {Proceedings of Machine Learning Research},
  pages        = {783--792},
  year         = {2019},
  url          = {http://proceedings.mlr.press/v97/brown19a.html}
}

@inproceedings{ziebart2008maximum,
  author       = {Brian D. Ziebart and
                  Andrew L. Maas and
                  J. Andrew Bagnell and
                  Anind K. Dey},
  title        = {Maximum Entropy Inverse Reinforcement Learning},
  booktitle    = {Proceedings of the Twenty-Third {AAAI} Conference on Artificial Intelligence,
                  {AAAI} 2008, Chicago, Illinois, USA, July 13-17, 2008},
  pages        = {1433--1438},
  year         = {2008},
  url          = {http://www.aaai.org/Library/AAAI/2008/aaai08-227.php}
}

@inproceedings{williams2017information,
  author       = {Grady Williams and
                  Nolan Wagener and
                  Brian Goldfain and
                  Paul Drews and
                  James M. Rehg and
                  Byron Boots and
                  Evangelos A. Theodorou},
  title        = {Information theoretic {MPC} for model-based reinforcement learning},
  booktitle    = {2017 {IEEE} International Conference on Robotics and Automation, {ICRA}
                  2017, Singapore, Singapore, May 29 - June 3, 2017},
  pages        = {1714--1721},
  year         = {2017},
  url          = {https://doi.org/10.1109/ICRA.2017.7989202},
  doi          = {10.1109/ICRA.2017.7989202}
}

@inproceedings{janner2022planning,
  author       = {Michael Janner and
                  Yilun Du and
                  Joshua B. Tenenbaum and
                  Sergey Levine},
  title        = {Planning with Diffusion for Flexible Behavior Synthesis},
  booktitle    = {International Conference on Machine Learning, {ICML} 2022, 17-23 July
                  2022, Baltimore, Maryland, {USA}},
  series       = {Proceedings of Machine Learning Research},
  pages        = {9902--9915},
  year         = {2022},
  url          = {https://proceedings.mlr.press/v162/janner22a.html}
}

@inproceedings{hansen2022tdmpc,
  author       = {Nicklas Hansen and
                  Hao Su and
                  Xiaolong Wang},
  title        = {Temporal Difference Learning for Model Predictive Control},
  booktitle    = {International Conference on Machine Learning, {ICML} 2022, 17-23 July
                  2022, Baltimore, Maryland, {USA}},
  series       = {Proceedings of Machine Learning Research},
  pages        = {8387--8406},
  year         = {2022},
  url          = {https://proceedings.mlr.press/v162/hansen22a.html}
}

@article{hafner2023dreamerv3,
  author       = {Danijar Hafner and
                  Jurgis Pasukonis and
                  Jimmy Ba and
                  Timothy P. Lillicrap},
  title        = {Mastering Diverse Domains through World Models},
  journal      = {CoRR},
  volume       = {abs/2301.04104},
  year         = {2023},
  url          = {https://doi.org/10.48550/arXiv.2301.04104},
  doi          = {10.48550/ARXIV.2301.04104},
  eprinttype   = {arXiv},
  eprint       = {2301.04104}
}

@article{argall2009survey,
  author       = {Brenna D. Argall and
                  Sonia Chernova and
                  Manuela M. Veloso and
                  Brett Browning},
  title        = {A survey of robot learning from demonstration},
  journal      = {Robotics Auton. Syst.},
  volume       = {57},
  number       = {5},
  pages        = {469--483},
  year         = {2009},
  url          = {https://doi.org/10.1016/j.robot.2008.10.024},
  doi          = {10.1016/J.ROBOT.2008.10.024}
}

@inproceedings{zhang2025flowpolicy,
  author       = {Qinglun Zhang and
                  Zhen Liu and
                  Haoqiang Fan and
                  Guanghui Liu and
                  Bing Zeng and
                  Shuaicheng Liu},
  title        = {FlowPolicy: Enabling Fast and Robust 3D Flow-Based Policy via Consistency
                  Flow Matching for Robot Manipulation},
  booktitle    = {Thirty-Ninth {AAAI} Conference on Artificial Intelligence, Thirty-Seventh
                  Conference on Innovative Applications of Artificial Intelligence,
                  Fifteenth Symposium on Educational Advances in Artificial Intelligence,
                  {AAAI} 2025, Philadelphia, PA, USA, February 25 - March 4, 2025},
  pages        = {14754--14762},
  year         = {2025},
  url          = {https://doi.org/10.1609/aaai.v39i14.33617},
  doi          = {10.1609/AAAI.V39I14.33617}
}

@inproceedings{lipman2022flow,
  author       = {Yaron Lipman and
                  Ricky T. Q. Chen and
                  Heli Ben{-}Hamu and
                  Maximilian Nickel and
                  Matthew Le},
  title        = {Flow Matching for Generative Modeling},
  booktitle    = {The Eleventh International Conference on Learning Representations,
                  {ICLR} 2023, Kigali, Rwanda, May 1-5, 2023},
  year         = {2023},
  url          = {https://openreview.net/forum?id=PqvMRDCJT9t}
}

@inproceedings{kostrikov2021offline,
  author       = {Ilya Kostrikov and
                  Ashvin Nair and
                  Sergey Levine},
  title        = {Offline Reinforcement Learning with Implicit Q-Learning},
  booktitle    = {The Tenth International Conference on Learning Representations, {ICLR}
                  2022, Virtual Event, April 25-29, 2022},
  year         = {2022},
  url          = {https://openreview.net/forum?id=68n2s9ZJWF8}
}

@article{qi2026gpc,
  author       = {Han Qi and
                  Haocheng Yin and
                  Aris Zhu and
                  Yilun Du and
                  Heng Yang},
  title        = {Inference-Time Enhancement of Generative Robot Policies via Predictive
                  World Modeling},
  journal      = {{IEEE} Robotics Autom. Lett.},
  volume       = {11},
  number       = {5},
  pages        = {5534--5541},
  year         = {2026},
  url          = {https://doi.org/10.1109/LRA.2026.3673995},
  doi          = {10.1109/LRA.2026.3673995}
}

@article{kim2026cosmospolicy,
  author       = {Moo Jin Kim and
                  Yihuai Gao and
                  Tsung{-}Yi Lin and
                  Yen{-}Chen Lin and
                  Yunhao Ge and
                  Grace Lam and
                  Percy Liang and
                  Shuran Song and
                  Ming{-}Yu Liu and
                  Chelsea Finn and
                  Jinwei Gu},
  title        = {Cosmos Policy: Fine-Tuning Video Models for Visuomotor Control and
                  Planning},
  journal      = {CoRR},
  volume       = {abs/2601.16163},
  year         = {2026},
  url          = {https://doi.org/10.48550/arXiv.2601.16163},
  doi          = {10.48550/ARXIV.2601.16163},
  eprinttype   = {arXiv},
  eprint       = {2601.16163}
}

@inproceedings{du2023video,
  author       = {Yilun Du and
                  Sherry Yang and
                  Bo Dai and
                  Hanjun Dai and
                  Ofir Nachum and
                  Josh Tenenbaum and
                  Dale Schuurmans and
                  Pieter Abbeel},
  title        = {Learning Universal Policies via Text-Guided Video Generation},
  booktitle    = {Advances in Neural Information Processing Systems 36: Annual Conference
                  on Neural Information Processing Systems 2023, NeurIPS 2023, New Orleans,
                  LA, USA, December 10 - 16, 2023},
  year         = {2023},
  url          = {http://papers.nips.cc/paper\_files/paper/2023/hash/1d5b9233ad716a43be5c0d3023cb82d0-Abstract-Conference.html}
}

@article{zhang2025worldinworld,
  author       = {Jiahan Zhang and
                  Muqing Jiang and
                  Nanru Dai and
                  Taiming Lu and
                  Arda Uzunoglu and
                  Shunchi Zhang and
                  Yana Wei and
                  Jiahao Wang and
                  Vishal M. Patel and
                  Paul Pu Liang and
                  Daniel Khashabi and
                  Cheng Peng and
                  Rama Chellappa and
                  Tianmin Shu and
                  Alan L. Yuille and
                  Yilun Du and
                  Jieneng Chen},
  title        = {World-in-World: World Models in a Closed-Loop World},
  journal      = {CoRR},
  volume       = {abs/2510.18135},
  year         = {2025},
  url          = {https://doi.org/10.48550/arXiv.2510.18135},
  doi          = {10.48550/ARXIV.2510.18135},
  eprinttype   = {arXiv},
  eprint       = {2510.18135}
}

@inproceedings{rajeswaran2018learning,
  author       = {Aravind Rajeswaran and
                  Vikash Kumar and
                  Abhishek Gupta and
                  Giulia Vezzani and
                  John Schulman and
                  Emanuel Todorov and
                  Sergey Levine},
  title        = {Learning Complex Dexterous Manipulation with Deep Reinforcement Learning
                  and Demonstrations},
  booktitle    = {Robotics: Science and Systems XIV, Carnegie Mellon University, Pittsburgh,
                  Pennsylvania, USA, June 26-30, 2018},
  year         = {2018},
  url          = {http://www.roboticsproceedings.org/rss14/p49.html},
  doi          = {10.15607/RSS.2018.XIV.049}
}

@inproceedings{yu2020metaworld,
  author       = {Tianhe Yu and
                  Deirdre Quillen and
                  Zhanpeng He and
                  Ryan Julian and
                  Karol Hausman and
                  Chelsea Finn and
                  Sergey Levine},
  title        = {Meta-World: {A} Benchmark and Evaluation for Multi-Task and Meta Reinforcement
                  Learning},
  booktitle    = {3rd Annual Conference on Robot Learning, CoRL 2019, Osaka, Japan,
                  October 30 - November 1, 2019, Proceedings},
  series       = {Proceedings of Machine Learning Research},
  pages        = {1094--1100},
  year         = {2019},
  url          = {http://proceedings.mlr.press/v100/yu20a.html}
}

@inproceedings{ross2011dagger,
  author       = {St{\'{e}}phane Ross and
                  Geoffrey J. Gordon and
                  Drew Bagnell},
  title        = {A Reduction of Imitation Learning and Structured Prediction to No-Regret
                  Online Learning},
  booktitle    = {Proceedings of the Fourteenth International Conference on Artificial
                  Intelligence and Statistics, {AISTATS} 2011, Fort Lauderdale, USA,
                  April 11-13, 2011},
  series       = {{JMLR} Proceedings},
  pages        = {627--635},
  year         = {2011},
  url          = {http://proceedings.mlr.press/v15/ross11a/ross11a.pdf}
}

@inproceedings{zhao2023learning,
  author       = {Tony Z. Zhao and
                  Vikash Kumar and
                  Sergey Levine and
                  Chelsea Finn},
  title        = {Learning Fine-Grained Bimanual Manipulation with Low-Cost Hardware},
  booktitle    = {Robotics: Science and Systems XIX, Daegu, Republic of Korea, July
                  10-14, 2023},
  year         = {2023},
  url          = {https://doi.org/10.15607/RSS.2023.XIX.016},
  doi          = {10.15607/RSS.2023.XIX.016}
}

@inproceedings{florence2022implicit,
  author       = {Pete Florence and
                  Corey Lynch and
                  Andy Zeng and
                  Oscar A. Ramirez and
                  Ayzaan Wahid and
                  Laura Downs and
                  Adrian Wong and
                  Johnny Lee and
                  Igor Mordatch and
                  Jonathan Tompson},
  title        = {Implicit Behavioral Cloning},
  booktitle    = {Conference on Robot Learning, 8-11 November 2021, London, {UK}},
  series       = {Proceedings of Machine Learning Research},
  pages        = {158--168},
  year         = {2021},
  url          = {https://proceedings.mlr.press/v164/florence22a.html}
}

@inproceedings{todorov2012mujoco,
  author       = {Emanuel Todorov and
                  Tom Erez and
                  Yuval Tassa},
  title        = {MuJoCo: {A} physics engine for model-based control},
  booktitle    = {2012 {IEEE/RSJ} International Conference on Intelligent Robots and
                  Systems, {IROS} 2012, Vilamoura, Algarve, Portugal, October 7-12,
                  2012},
  pages        = {5026--5033},
  year         = {2012},
  url          = {https://doi.org/10.1109/IROS.2012.6386109},
  doi          = {10.1109/IROS.2012.6386109}
}

@inproceedings{finn2017deep,
  author       = {Chelsea Finn and
                  Sergey Levine},
  title        = {Deep visual foresight for planning robot motion},
  booktitle    = {2017 {IEEE} International Conference on Robotics and Automation, {ICRA}
                  2017, Singapore, Singapore, May 29 - June 3, 2017},
  pages        = {2786--2793},
  year         = {2017},
  url          = {https://doi.org/10.1109/ICRA.2017.7989324},
  doi          = {10.1109/ICRA.2017.7989324}
}

@inproceedings{ho2020denoising,
  author       = {Jonathan Ho and
                  Ajay Jain and
                  Pieter Abbeel},
  title        = {Denoising Diffusion Probabilistic Models},
  booktitle    = {Advances in Neural Information Processing Systems 33: Annual Conference
                  on Neural Information Processing Systems 2020, NeurIPS 2020, December
                  6-12, 2020, virtual},
  year         = {2020},
  url          = {https://proceedings.neurips.cc/paper/2020/hash/4c5bcfec8584af0d967f1ab10179ca4b-Abstract.html}
}

@inproceedings{song2020denoising,
  author       = {Jiaming Song and
                  Chenlin Meng and
                  Stefano Ermon},
  title        = {Denoising Diffusion Implicit Models},
  booktitle    = {9th International Conference on Learning Representations, {ICLR} 2021,
                  Virtual Event, Austria, May 3-7, 2021},
  year         = {2021},
  url          = {https://openreview.net/forum?id=St1giarCHLP}
}

@inproceedings{song2023consistency,
  author       = {Yang Song and
                  Prafulla Dhariwal and
                  Mark Chen and
                  Ilya Sutskever},
  title        = {Consistency Models},
  booktitle    = {International Conference on Machine Learning, {ICML} 2023, 23-29 July
                  2023, Honolulu, Hawaii, {USA}},
  series       = {Proceedings of Machine Learning Research},
  pages        = {32211--32252},
  year         = {2023},
  url          = {https://proceedings.mlr.press/v202/song23a.html}
}

@article{deboer2005tutorial,
  author       = {Pieter{-}Tjerk de Boer and
                  Dirk P. Kroese and
                  Shie Mannor and
                  Reuven Y. Rubinstein},
  title        = {A Tutorial on the Cross-Entropy Method},
  journal      = {Ann. Oper. Res.},
  volume       = {134},
  number       = {1},
  pages        = {19--67},
  year         = {2005},
  url          = {https://doi.org/10.1007/s10479-005-5724-z},
  doi          = {10.1007/S10479-005-5724-Z}
}

@article{wang2024qwen2vl,
  author       = {Peng Wang and
                  Shuai Bai and
                  Sinan Tan and
                  Shijie Wang and
                  Zhihao Fan and
                  Jinze Bai and
                  Keqin Chen and
                  Xuejing Liu and
                  Jialin Wang and
                  Wenbin Ge and
                  Yang Fan and
                  Kai Dang and
                  Mengfei Du and
                  Xuancheng Ren and
                  Rui Men and
                  Dayiheng Liu and
                  Chang Zhou and
                  Jingren Zhou and
                  Junyang Lin},
  title        = {Qwen2-VL: Enhancing Vision-Language Model's Perception of the
                  World at Any Resolution},
  journal      = {CoRR},
  volume       = {abs/2409.12191},
  year         = {2024},
  url          = {https://doi.org/10.48550/arXiv.2409.12191},
  doi          = {10.48550/ARXIV.2409.12191},
  eprinttype   = {arXiv},
  eprint       = {2409.12191}
}

@inproceedings{shridhar2023perceiver,
  author       = {Mohit Shridhar and
                  Lucas Manuelli and
                  Dieter Fox},
  title        = {Perceiver-Actor: {A} Multi-Task Transformer for Robotic Manipulation},
  booktitle    = {Conference on Robot Learning, CoRL 2022, 14-18 December 2022, Auckland,
                  New Zealand},
  series       = {Proceedings of Machine Learning Research},
  pages        = {785--799},
  year         = {2022},
  url          = {https://proceedings.mlr.press/v205/shridhar23a.html}
}

@inproceedings{gervet2023act3d,
  author       = {Th{\'{e}}ophile Gervet and
                  Zhou Xian and
                  Nikolaos Gkanatsios and
                  Katerina Fragkiadaki},
  title        = {Act3D: 3D Feature Field Transformers for Multi-Task Robotic Manipulation},
  booktitle    = {Conference on Robot Learning, CoRL 2023, 6-9 November 2023, Atlanta,
                  GA, {USA}},
  series       = {Proceedings of Machine Learning Research},
  pages        = {3949--3965},
  year         = {2023},
  url          = {https://proceedings.mlr.press/v229/gervet23a.html}
}

@inproceedings{goyal2023rvt,
  author       = {Ankit Goyal and
                  Jie Xu and
                  Yijie Guo and
                  Valts Blukis and
                  Yu{-}Wei Chao and
                  Dieter Fox},
  title        = {{RVT:} Robotic View Transformer for 3D Object Manipulation},
  booktitle    = {Conference on Robot Learning, CoRL 2023, 6-9 November 2023, Atlanta,
                  GA, {USA}},
  series       = {Proceedings of Machine Learning Research},
  pages        = {694--710},
  year         = {2023},
  url          = {https://proceedings.mlr.press/v229/goyal23a.html}
}
}


\appendix

\section{Benchmark tasks and simulation environments}

\begin{figure}[h]
    \centering
    \setlength{\tabcolsep}{3pt}
    \begin{tabular}{ccc}
        \includegraphics[width=0.31\linewidth]{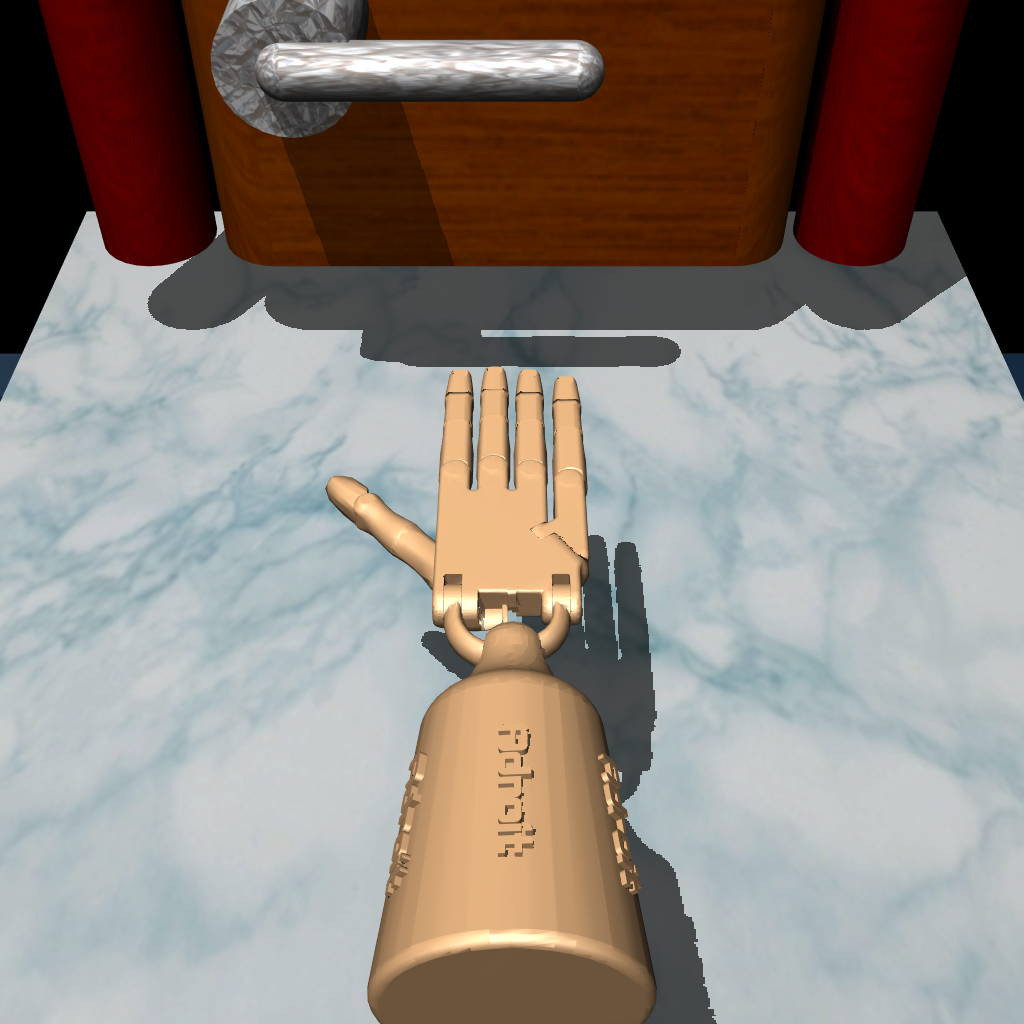} &
        \includegraphics[width=0.31\linewidth]{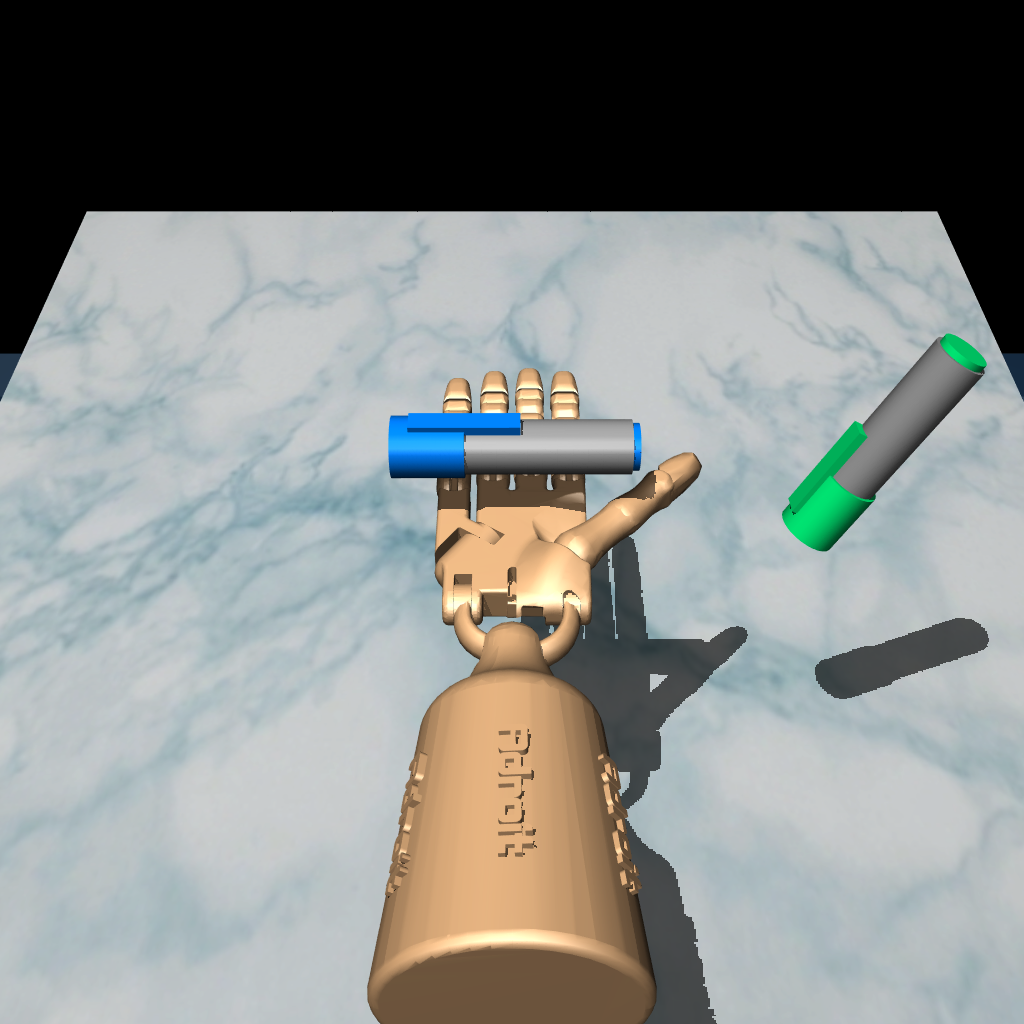} &
        \includegraphics[width=0.31\linewidth]{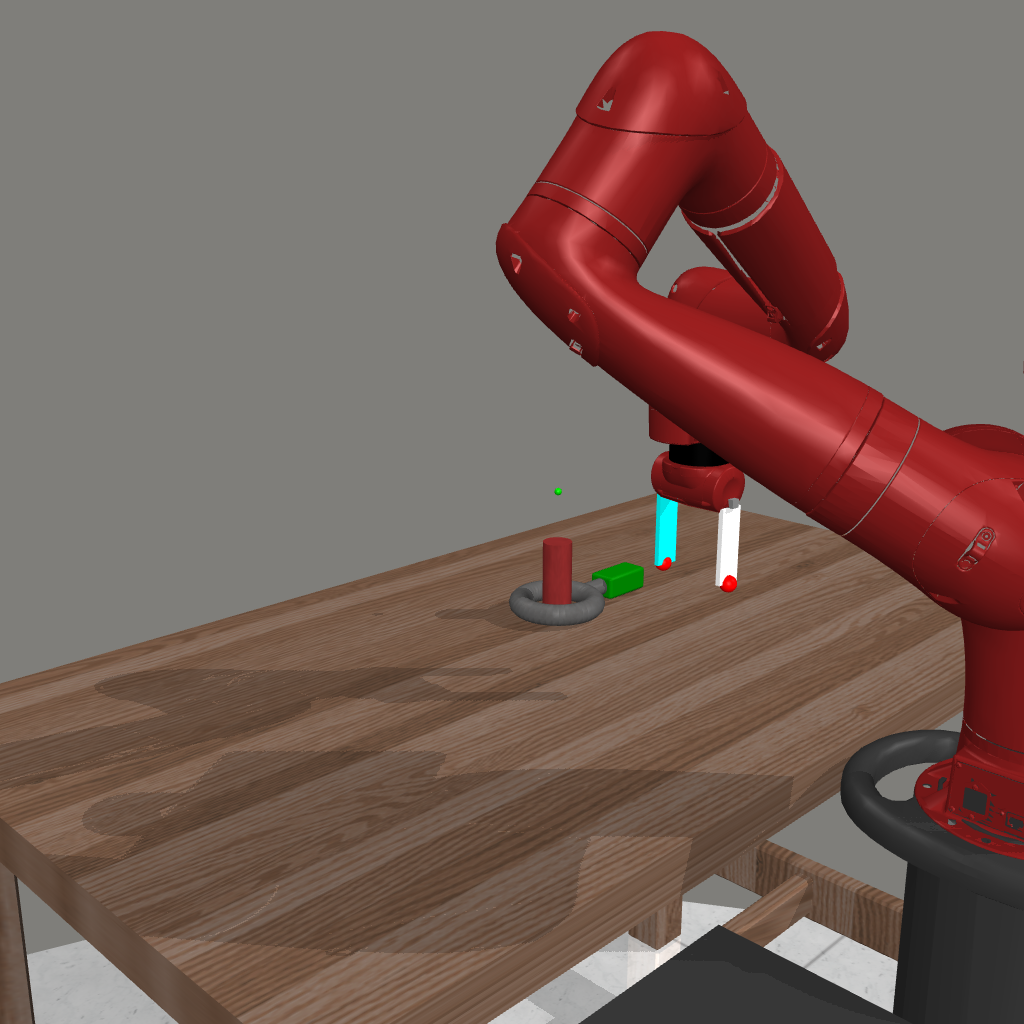} \\[0.5ex] 
        \includegraphics[width=0.31\linewidth]{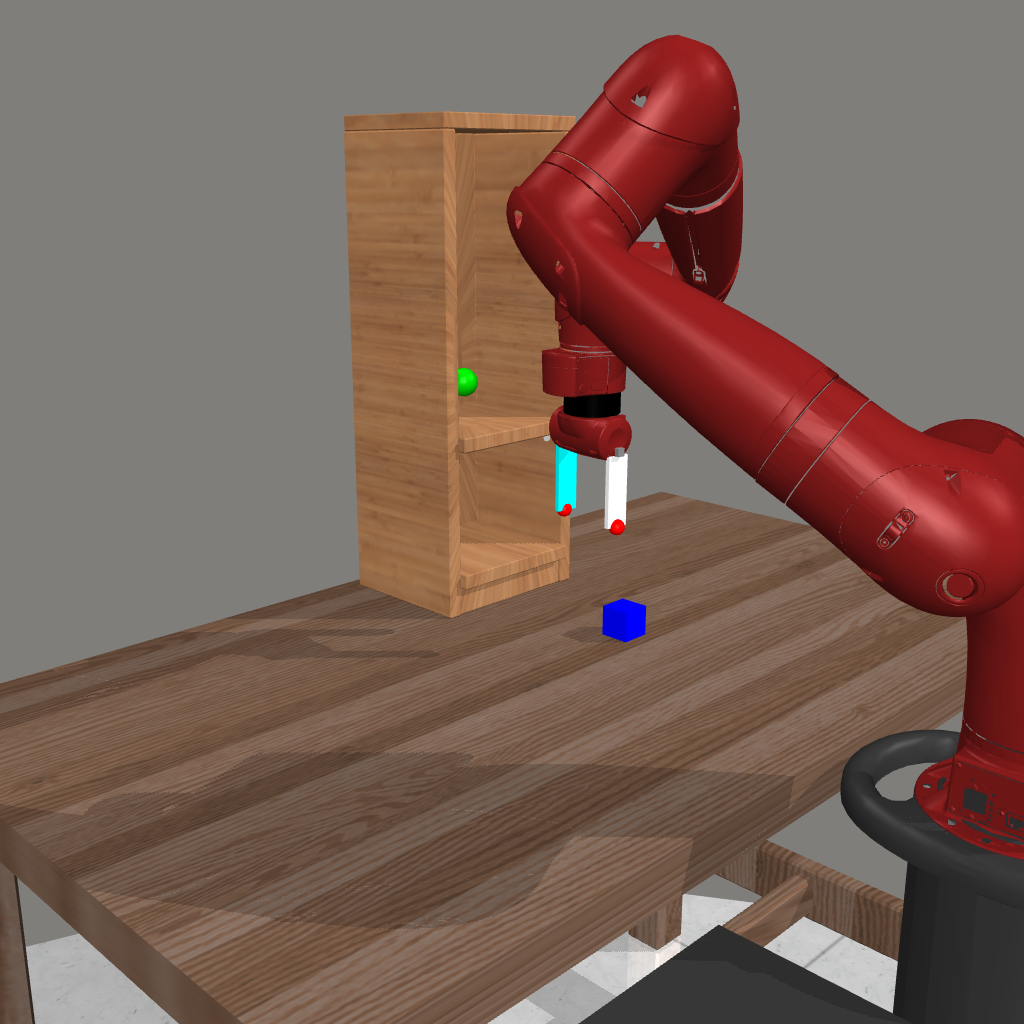} &
        \includegraphics[width=0.31\linewidth]{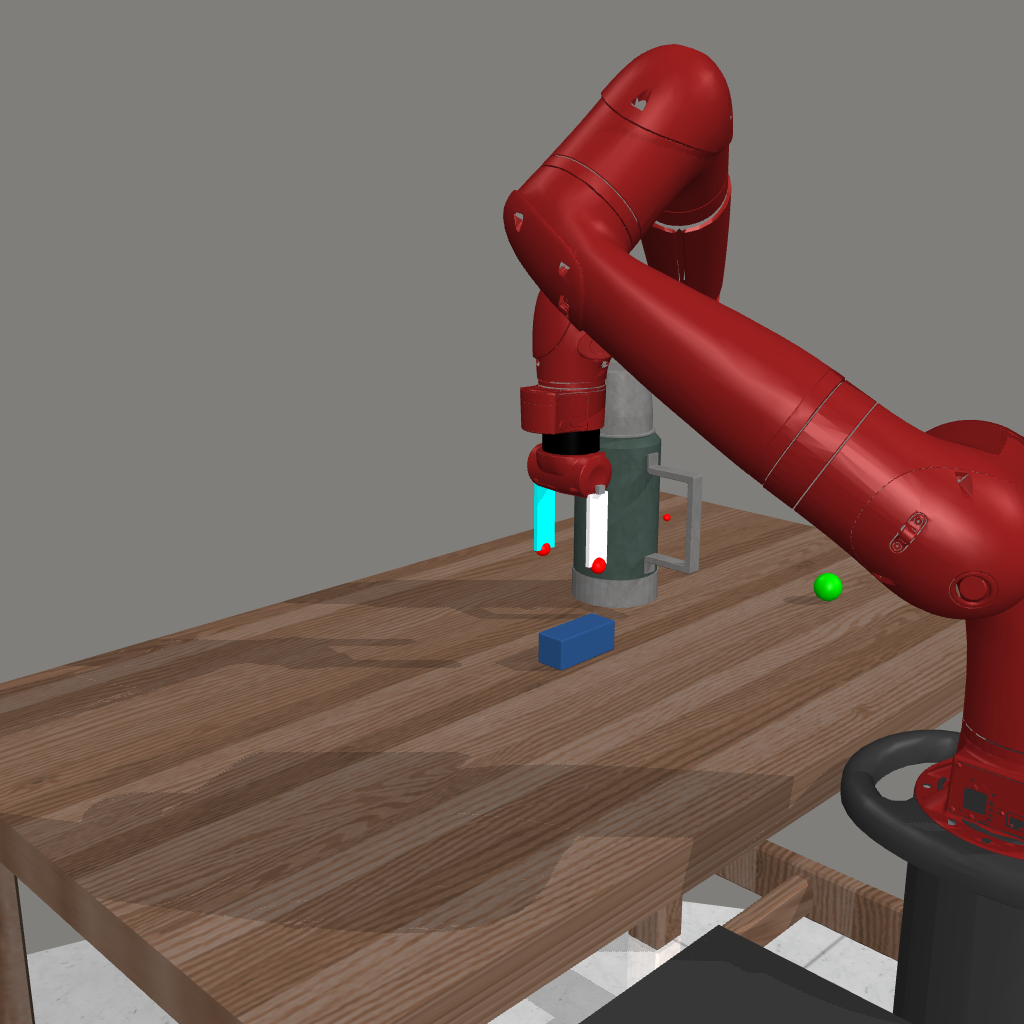} &
        \includegraphics[width=0.31\linewidth]{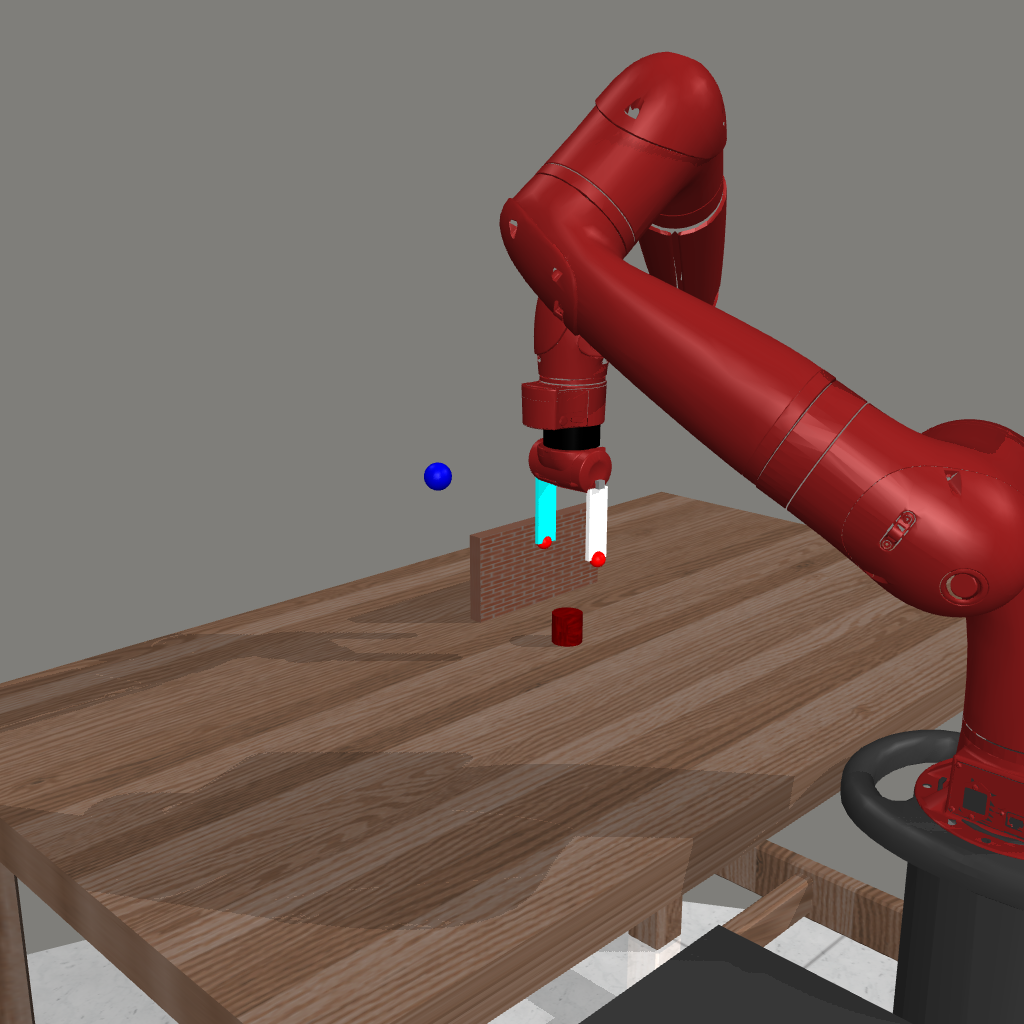}
    \end{tabular}
    \vspace{-0.5em} 
    \caption{
    \textbf{Representative rendered observations from the evaluated Adroit and MetaWorld simulation tasks.} 
    \textbf{Top row (left to right):} Door, Pen, Disassemble. \textbf{Bottom row (left to right):} Shelf-Place, Stick-Pull, Pick-Place-Wall. 
    Door and Pen use the Adroit dexterous-hand embodiment, while the remaining tasks use the MetaWorld gripper embodiment.
    }
    \label{fig:appendix_task_overview}
\end{figure}

Figure~\ref{fig:appendix_task_overview} shows representative rendered observations
from the six evaluated simulation tasks. We evaluate two Adroit dexterous-hand tasks,
Door and Pen, and four MetaWorld gripper manipulation tasks: Disassemble,
Shelf-Place, Stick-Pull, and Pick-Place-Wall.

\section{Experimental setup}

\subsection{Datasets and evaluation protocol}

We use the six simulation tasks reported in the main paper: Door and Pen from
Adroit, and Disassemble, Shelf-Place, Stick-Pull, and Pick-Place-Wall from
MetaWorld. To isolate the effect of using imperfect trajectories for prediction
and evaluation, all base policies are trained with the same 10 expert
demonstration episodes per task. DALI-R additionally uses 1000 mixed-quality
trajectories per task to train only the Latent World Model and Task Completion
Scorer. These mixed trajectories are generated by policy rollouts and by adding
drift noise to actions from the expert policy that provides the expert
demonstrations. They are never used as imitation targets for the Base 3D Policy.

Trajectories are partitioned according to the terminal task-completion signal
returned by the simulator. Successful expert and imperfect-success trajectories
are assigned terminal label 1, while failed trajectories are assigned terminal
label 0. For the offline real-world diagnostic, we split released 3D Diffusion
Policy trajectories at the trajectory level to avoid frame-level leakage and
report held-out latent transition MSE rather than closed-loop real-robot
performance.

\subsection{Implementation details}

All policies use horizon 16, 2 observation steps, and 8 action execution
steps, following the DP3 configuration. The Task Completion Scorer is trained with AdamW using learning rate $1\times10^{-4}$, batch size 128, weight decay $1\times10^{-3}$, and 30 epochs, with binary cross entropy on discounted terminal task-completion targets.

The Latent World Model predicts chunk-level latent residual transitions
conditioned on the current latent state and an action chunk. It is implemented as a lightweight Transformer encoder with hidden dimension 256, 3 layers, 4 attention heads, and action-chunk length 8. The world model is trained with AdamW using learning rate $3\times10^{-4}$, weight decay $1\times10^{-4}$, batch size 512, and 100 epochs, with mean-squared error loss.

At inference time, DALI-R samples $N_{\mathrm{cand}}=32$ candidate action chunks from the frozen Base 3D Policy. Candidate diversity is induced by a fixed stochastic point-cloud perturbation setting that is kept unchanged across tasks and methods. Each candidate is evaluated by one-step latent world-model prediction followed by Task Completion Scorer evaluation. All experiments are conducted on a single NVIDIA RTX 3090 GPU with 24GB memory.

\section{Additional experimental details}
\subsection{Additional latency details}
Table~\ref{tab:appendix_latency} reports the exact latency and FPS values
corresponding to Figure~3(a) in the main paper. The 2D Video + VLM pipeline is
included only as a simulated latency reference rather than a trained task
baseline.

\begin{table}[H]
    \centering
    \small
    \setlength{\tabcolsep}{5pt}
    \caption{
    Exact inference latency and FPS values under different numbers of candidate
    action chunks. Latency is measured in milliseconds per decision.
    }
    \label{tab:appendix_latency}
    \begin{tabular}{c|cc|cc|cc}
        \toprule
        \multirow{2}{*}{$N_{\mathrm{cand}}$}
        & \multicolumn{2}{c|}{2D Video + VLM}
        & \multicolumn{2}{c|}{3D DP + Ours}
        & \multicolumn{2}{c}{3D FM + Ours} \\
        \cmidrule(lr){2-3}
        \cmidrule(lr){4-5}
        \cmidrule(lr){6-7}
        & Lat. $\downarrow$ & FPS $\uparrow$
        & Lat. $\downarrow$ & FPS $\uparrow$
        & Lat. $\downarrow$ & FPS $\uparrow$ \\
        \midrule
        1  & 335.23   & 2.98  & 70.72  & 14.14 & 7.83  & 127.63 \\
        4  & 1414.91  & 0.71  & 83.17  & 12.02 & 10.64 & 94.03  \\
        8  & 2721.85  & 0.37  & 87.95  & 11.37 & 11.09 & 90.14  \\
        16 & 5451.23  & 0.18  & 93.32  & 10.72 & 11.82 & 84.58  \\
        32 & 10948.59 & 0.09  & 108.92 & 9.18  & 13.62 & 73.44  \\
        \bottomrule
    \end{tabular}
\end{table}

\subsection{Inference-time algorithms}

All inference-time action-improvement methods in Table~\ref{tab:action_improvement}
use the same frozen 3D FM base policy, Latent World Model, and Task Completion
Scorer. They differ only in how candidate action chunks are generated.

\paragraph{Observation-space methods.}
DALI-R uses stochastic point dropout to generate perturbed point-cloud
observations, queries the frozen Base 3D Policy under these observations, and
selects the highest-scoring policy-generated action chunk after latent
world-model prediction and task-completion scoring. The standard observation
noise baseline follows the same procedure but replaces point dropout with
unstructured Gaussian perturbations. In both cases, one unperturbed observation
is retained as a candidate.

\paragraph{Action-space methods.}
Random shooting samples noisy action chunks around the base policy proposal and selects the highest-scoring one. CEM iteratively updates a Gaussian action proposal distribution using top-scoring elite samples. MPPI forms a
score-weighted average of noisy action samples using a softmax over scorer
values. Score-gradient ascent directly optimizes the action chunk by
backpropagating through the frozen Latent World Model and Task Completion Scorer with an action-deviation penalty. For conservative execution, action-space methods keep the gripper dimension unchanged and accept modified actions only when they improve the scorer value over the base policy proposal.

Algorithm~\ref{alg:latent_reranking} summarizes the default DALI-R inference procedure.

\begin{algorithm}[H]
\caption{Inference-time latent reranking with stochastic point dropout}
\label{alg:latent_reranking}
\begin{algorithmic}
\Require Frozen Base 3D Policy $\pi_\theta$, latent encoders $E^o_\phi, E^q_\eta$, Latent World Model $W_\psi$, Task Completion Scorer $C_\omega$
\Require Number of candidates $N_{\mathrm{cand}}$, action horizon $k$
\While{task is not completed}
    \State Receive current point-cloud observation $o_t$ and robot state $q_t$
    \State Encode latent state $s_t \gets [E_\eta^q(q_t), E_\phi^o(o_t)]$
    \For{$i=1$ to $N_{\mathrm{cand}}$ \textbf{in parallel}}
        \State Apply stochastic point dropout: $\tilde{o}_t^{(i)} \gets \mathrm{Dropout}(o_t)$
        \State Generate candidate action chunk:
        $\mathbf{a}_{t:t+k}^{(i)} \sim \pi_\theta(\cdot \mid \tilde{o}_t^{(i)}, q_t)$
        \State Imagine future latent:
        $\hat{s}_{t+k}^{(i)} \gets s_t + W_\psi(s_t, \mathbf{a}_{t:t+k}^{(i)})$
        \State Score imagined latent:
        $p^{(i)} \gets \sigma(C_\omega(\hat{s}_{t+k}^{(i)}))$
    \EndFor
    \State Select $i^\star \gets \arg\max_i p^{(i)}$
    \State Execute $\mathbf{a}_{t:t+k}^{(i^\star)}$ according to the receding-horizon schedule
\EndWhile
\end{algorithmic}
\end{algorithm}

\section{Compute resources}

All experiments were conducted on a single NVIDIA RTX 3090 GPU with 24GB memory.
Training and evaluation were implemented in PyTorch. Latency measurements were
performed on the same hardware after GPU warm-up. The reported latency values
measure milliseconds per decision for a single closed-loop control step.

\section{Existing assets and licenses}

We use existing simulation environments, codebases, models, and released trajectories only for research evaluation and offline diagnostics; no existing asset is redistributed with this submission.

The Adroit and MetaWorld simulation tasks are credited in the main paper. MetaWorld is distributed under the MIT License. The Adroit environments are used through Gymnasium-Robotics/D4RL-style Adroit code; Gymnasium-Robotics is distributed under the MIT License, and D4RL code is distributed under the Apache-2.0 License, with datasets licensed under CC BY 4.0 unless otherwise noted. MuJoCo, the physics simulator used by these benchmarks, is open source and distributed under the Apache-2.0 License.

We build on and cite 3D Diffusion Policy (DP3); its public repository, including the released real-robot trajectories used for the offline diagnostic, is distributed under the MIT License unless separate terms are specified. The 3D flow-matching policy used in our experiments is our own implementation; FlowPolicy is cited as related prior work and not used as a code or model asset. The Qwen2-VL-2B model used only as a latency reference for the 2D Video + VLM pipeline is released under the Apache-2.0 License. We do not release new datasets, pretrained models, or benchmark assets as part of this submission.


\end{document}